\documentclass[a4paper,conference]{IEEEtran}
\ifCLASSINFOpdf
\else
\fi

\usepackage[table]{xcolor}
\usepackage{graphicx}
\usepackage{multirow,multicol}
\usepackage{siunitx}
\usepackage{booktabs}       
\usepackage{tikz}
\usepackage{amssymb}

\makeatletter
\global\let\oriCT@@do@color\CT@@do@color

\begin{document}
\title{Learning-based Monocular 3D Reconstruction of Birds: A Contemporary Survey}

\author{\IEEEauthorblockN{
Seyed Mojtaba Marvasti-Zadeh\IEEEauthorrefmark{1}\textsuperscript{\textsection},
Mohammad N.S. Jahromi\IEEEauthorrefmark{2}\textsuperscript{\textsection}, 
Javad Khaghani\IEEEauthorrefmark{10},
Devin Goodsman\IEEEauthorrefmark{5}, \\
Nilanjan Ray\IEEEauthorrefmark{3}, and
Nadir Erbilgin\IEEEauthorrefmark{1}
}

%

\IEEEauthorblockA{
\IEEEauthorrefmark{1}Department of Renewable Resources, University of Alberta, Canada \\
\IEEEauthorrefmark{2}Research and Technology Department, Demant A/S, Copenhagen, Denmark \\
\IEEEauthorrefmark{10}Department of Electrical and Computer Engineering, University of Alberta, Canada \\
\IEEEauthorrefmark{5}Canadian Forest Service, Natural Resources Canada, Edmonton, Alberta \\
\IEEEauthorrefmark{3}Department of Computing Science, University of Alberta, Canada 
}
Emails: \{seyedmoj, khaghani, nray1, erbilgin\}@ualberta.ca, mosj@demant.com, devin.goodsman@nrcan-rncan.gc.ca
} 
\maketitle
\begingroup\renewcommand\thefootnote{\textsection}
\footnotetext{Equal contribution.}
\endgroup

\maketitle

\begin{abstract}
In nature, the collective behavior of animals, such as flying birds is dominated by the interactions between individuals of the same species. However, the study of such behavior among the bird species is a complex process that humans cannot perform using conventional visual observational techniques such as focal sampling in nature. For social animals such as birds, the mechanism of group formation can help ecologists understand the relationship between social cues and their visual characteristics over time (e.g., pose and shape). But, recovering the varying pose and shapes of flying birds is a highly challenging problem. A widely-adopted solution to tackle this bottleneck is to extract the pose and shape information from 2D image to 3D correspondence. Recent advances in 3D vision have led to a number of impressive works on the 3D shape and pose estimation, each with different pros and cons. 
To the best of our knowledge, this work is the first attempt to provide an overview of recent advances in 3D bird reconstruction based on monocular vision, give both computer vision and biology researchers an overview of existing approaches, and compare their characteristics. 
\end{abstract}


%
\IEEEpeerreviewmaketitle


\section{Introduction}
\noindent Understanding the collective behavior of animals based on the interaction between individuals is an important task in many disciplines such as evolutionary biology, computational biology, and neuroscience~\cite{vidal2021perspectives,ref2}. An excellent example of such behavior influencing social evolution can be observed for different bird species. For instance, changes in the visual pose and shape of flying birds over time help researchers understand and explain various aspects of social communication among birds through visual mechanisms~\cite{ref2}. In general, capturing such visual observation is performed by experts in the field over a continuous time period, often resulting in inaccurate measurements and observer bias. Consequently, to exploit the visual characteristics of birds, a reliable automated capture is required to estimate the birds' visual features such as pose and shape, respectively. However, estimating the pose and shape of birds in the wild is complicated for several reasons, such as the wide range of variations in shape and appearance~\cite{ref26}. \\
\begin{figure}[!t]
\centering
\includegraphics[width=.75\linewidth]{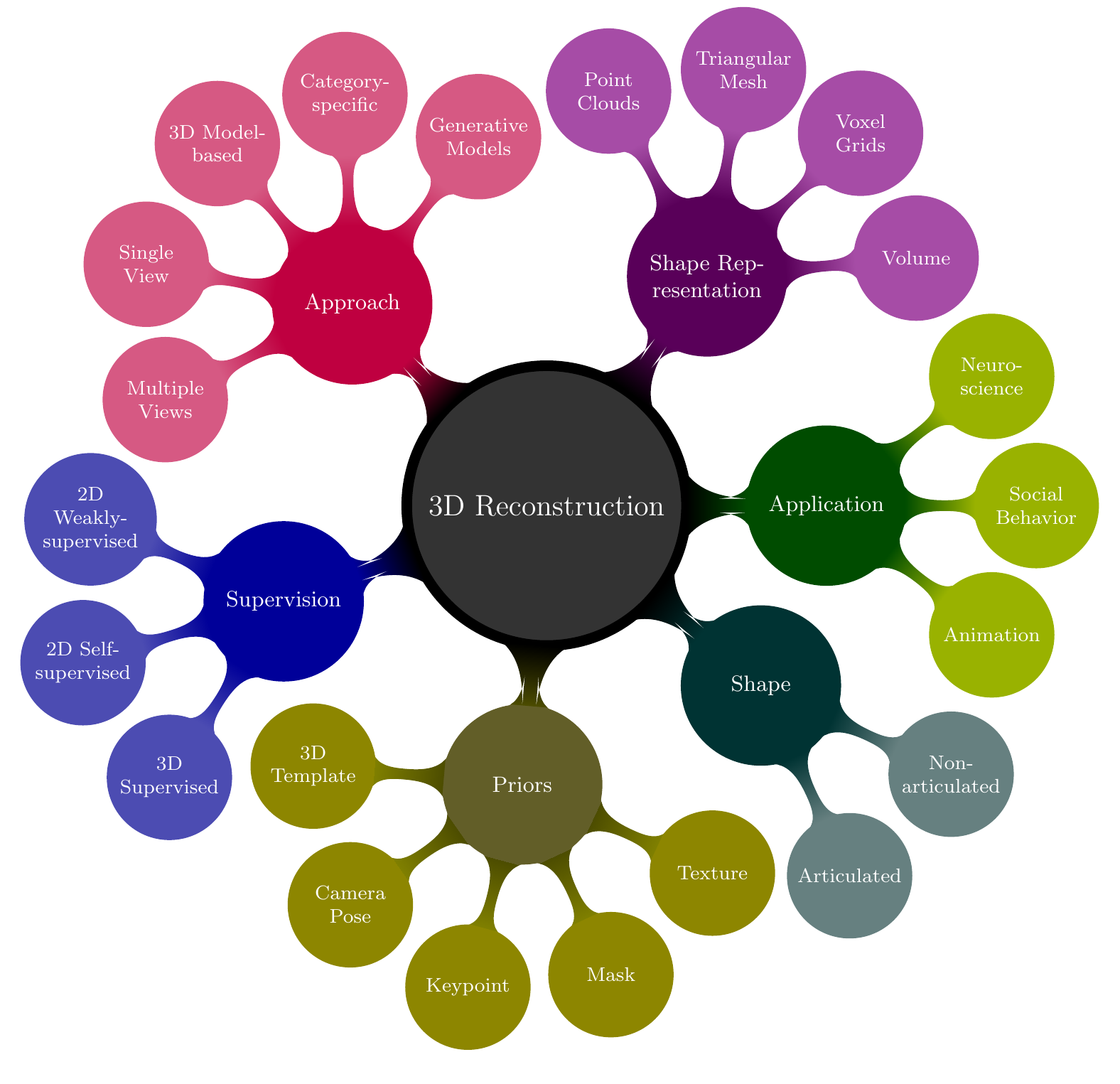}
\vspace{-.3cm}
\caption{Brief overview of 3D reconstruction methods for birds. This paper explores single-view, mesh-based approaches for articulated/non-articulated shapes utilizing various priors with 2D weak/self-supervisions.}
\vspace{-.55cm}
\label{fig:overview}
\end{figure}
\indent Due to various technical challenges resulting from collecting multi-view 2D images or labeling 3D models in the wild, a new class of methods has emerged in 3D computer vision that attempts to reconstruct 3D objects from monocular vision (i.e., single-view images). 
Recent advances in 3D computer vision and deep learning have provided a variety of 3D reconstruction approaches from a 2D image collection or video sequence to estimate the pose and shape of a general object category. The proposed approaches are mainly benchmarked on two well-suited datasets, namely the Caltech-UCSD Birds (CUB \num{200}-\num{2007})~\cite{ref6}~(\num{6033} images of \num{200} bird species) and the extended version CUB \num{200}-\num{2011}~\cite{ref7}~(\num{11788} images of \num{200} bird species) image collections tagged with keypoints, bounding boxes, coarse segmentation, and attribute labels.
The main idea behind these approaches is to render 3D meshes from 2D image representation without the need to have synthetic data/multi-view images or 3D supervision~\cite{pavllo2021learning}.  However, the monocular 3D reconstruction of general object categories is an extremely challenging task due to the inherent ambiguities arising from various sources of geometric variability (e.g., different camera poses and shapes).\\ 
\indent The mesh-based approach is preferred among the existing approaches (e.g., generative-based~\cite{pavllo2021learning} or model-based~\cite{biggs2020left} (see Fig.~\ref{fig:overview})) because of its high shape expressivity (e.g., finer shape details). However, mesh-based methods depend on various types of supervision, such as initial 3D shapes, 2D semantic keypoints, and foreground masks. These priors or features extracted from 2D images serve as supervisory signals during learning when ground truth 3D shape is not available. To recover 3D structures from an image/video, these signals help the learning frameworks encode the 3D shape, camera pose, and texture associated with them. \\
\indent Even though these learning-based methods have delivered promising results, tedious and careful 2D annotations are required that practically correspond to a handful of categories (i.e., limited generalization). Besides, the effectiveness of these methods can be dramatically affected by various issues such as large shape variations, invisible keypoints, and occlusion~\cite{ref36}. Furthermore, achieving impressive results requires acquiring large image collection and handling ambiguities associated with 2D observations across instances and viewpoints. Video sequences can be used as an alternative to image collection which is more convenient to deal with representing single/multi-view of the 3D shape of the same object. Moreover, utilizing videos can provide fine-grained information like optical flow to process motion beyond just instance keypoints or semantic parts~\cite{ref33}. There are still challenges to be overcome, such as reducing supervisory signals, providing generalization to other species, and dealing with data requirements and issues in real-world scenarios. To the best of our knowledge, no contribution has been made so far that has thoroughly reviewed the monocular 3D reconstruction of birds and its challenges from in-the-wild images/videos. 
Hence, this paper focuses on reviewing the learning-based image/video input approaches for 3D reconstruction of birds from 2D images and comparing their features, advantages, and limitations.
\begin{table*}[!h]
    \rowcolors{2}{}{gray!10}
    \caption{Comparison of learning-based monocular 3D birds reconstruction.}
    \vspace{-.2cm}
    \label{tab:comparison}
    \centering
    \resizebox{\textwidth}{!}{
    \begin{tabular}{c c | c | c c c c c | c c c c | c }
        \toprule
            &  &  & \multicolumn{5}{c|}{Input} & \multicolumn{4}{c|}{Output} &    \\ 
           
            \global\let\CT@@do@color\relax & \global\let\CT@@do@color\oriCT@@do@color Method & \global\let\CT@@do@color\oriCT@@do@color Type of 2D Supervision & Keypoint & Mask & Camera pose & Template & \multicolumn{1}{c|}{Optical flow} & 3D Shape & Camera Pose & Texture & Deformation & Backbone  \\ 
           
        \midrule
        
            
            \global\let\CT@@do@color\relax  \multirow{10}{*}{\rotatebox[origin=c]{90}{Image}} & \global\let\CT@@do@color\oriCT@@do@color CMR \cite{ref5} & Weakly-supervised & \checkmark & \checkmark & SfM & initial shape & $\times$ & \checkmark & \checkmark & UV-flow & $\times$ & ResNet-18 \\
            
            \global\let\CT@@do@color\relax & \global\let\CT@@do@color\oriCT@@do@color CSM \cite{ref29} & Weakly-supervised & $\times$ & \checkmark & $\times$ & \checkmark & $\times$ & $\times$ & \checkmark & $\times$ & $\times$ & ResNet-18 \\
            
            \global\let\CT@@do@color\relax & \global\let\CT@@do@color\oriCT@@do@color U-CMR \cite{ref8} & Weakly-supervised & $\times$ & \checkmark & $\times$ & \checkmark & $\times$ & \checkmark & \checkmark & \checkmark & $\times$ & ResNet-18 \\
            
            \global\let\CT@@do@color\relax & \global\let\CT@@do@color\oriCT@@do@color UMR \cite{ref9} & Self-supervised & $\times$ & \checkmark & $\times$ & $\times$ & $\times$ & \checkmark & \checkmark & UV-flow & $\times$ & ResNet-18 \\ 
            
            \global\let\CT@@do@color\relax & \global\let\CT@@do@color\oriCT@@do@color WCMR \cite{ref33} & Weakly-supervised & \checkmark & \checkmark & $\times$ & initial shape & $\times$ & \checkmark & \checkmark & UV-flow & \checkmark & ResNet-18 \\ 
            
            \global\let\CT@@do@color\relax & \global\let\CT@@do@color\oriCT@@do@color ACSM \cite{ref3} & Weakly- or Self-supervised & $\times$ & \checkmark & $\times$ & \checkmark & $\times$ & $\times$ & \checkmark & $\times$ & \checkmark & ResNet-18 \\ 
            
            \global\let\CT@@do@color\relax & \global\let\CT@@do@color\oriCT@@do@color IMR \cite{ref36} & Weakly- or Self-supervised & $\times$ & \checkmark & $\times$ & initial shape & $\times$ & \checkmark & \checkmark & UV-flow & $\times$ & ResNet-18 \\ 
            
            \global\let\CT@@do@color\relax & \global\let\CT@@do@color\oriCT@@do@color TTP \cite{ref28} & Weakly- or Self-supervised & $\times$ & \checkmark & $\times$ & initial shape & $\times$ & $\times$ & \checkmark & \checkmark & \checkmark & ResNet-18 \\ 
            
            \global\let\CT@@do@color\relax & \global\let\CT@@do@color\oriCT@@do@color ACMR \cite{ref25} & Weakly-supervised & \checkmark & \checkmark & $\times$ & $\times$ & $\times$ & \checkmark & \checkmark & UV-flow & \checkmark & ResNet-18 \\ 
            
            \global\let\CT@@do@color\relax & \global\let\CT@@do@color\oriCT@@do@color AVES \cite{ref4} & Supervised & \checkmark & \checkmark & \checkmark & articulated shape & $\times$ & \checkmark & \checkmark & $\times$ & \checkmark & ResNet-50 \\ 
            
        \midrule            
        
            \global\let\CT@@do@color\relax \multirow{4}{*}{\rotatebox[origin=c]{90}{Video}} & \global\let\CT@@do@color\oriCT@@do@color ACMR-vid \cite{ref25} & Self-supervised & $\times$ & \checkmark & $\times$ & $\times$ & $\times$ & \checkmark & \checkmark & UV-flow & \checkmark & ResNet-50  \\ 
            
            \global\let\CT@@do@color\relax & \global\let\CT@@do@color\oriCT@@do@color ACM \cite{ref31} & Weakly-supervised & \checkmark & \checkmark & $\times$ & \checkmark & \checkmark & \checkmark & \checkmark & \checkmark & \checkmark & ResNet-18 \\ 
            
            \global\let\CT@@do@color\relax & \global\let\CT@@do@color\oriCT@@do@color LASR \cite{ref26} & Self-supervised & $\times$ & \checkmark & $\times$ & $\times$ & \checkmark & \checkmark & \checkmark & \checkmark & \checkmark & ResNet-18  \\ 
            
            \global\let\CT@@do@color\relax & \global\let\CT@@do@color\oriCT@@do@color DOVE \cite{ref34} & Self-supervised & $\times$ & \checkmark & $\times$ & $\times$ & \checkmark & \checkmark & \checkmark & \checkmark & \checkmark & Customized CNN \\
            
        \bottomrule
    \end{tabular}
    }
    \vspace{-.3cm}
\end{table*}
\section{Monocular 3D Reconstruction: Taxonomy}
\noindent In this section, the approaches for reconstructing 3D models of birds using monocular images/videos are discussed. Generally speaking, the objectives of these approaches are to reduce 2D supervision requirements and capture fine details through learning. Accordingly, the state-of-the-art approaches inferring the 3D structure of birds are compared in Table~\ref{tab:comparison}.

\subsection{Image Collection-based Approaches}
A deeper understanding of 3D structure in images is contingent upon the global and local relationships between 2D percepts and 3D concepts. Hence, developing a computational model of birds can provide insight into their 3D structures. To learn this model, the well-known \textit{category-specific mesh reconstruction} (CMR) \cite{ref5} leverages 2D image collections annotated by foreground masks and semantic keypoint labels to recover 3D bird shape \& texture and camera pose from a single image at inference. The CMR trains a ResNet-18-based encoder with three modules to efficiently predict deformable 3D mesh representations, assuming mirror-symmetry constraints and smooth surfaces. It explicitly learns to assign semantic keypoints of birds (such as legs \& beaks) to a 3D mesh, which is enforced to be consistent with the foreground mask and natural world assumptions. In addition, the CMR relaxes the particular instance texturing to the fixed UV mapping of a mean shape with consistent semantic meaning.
Although the CMR is able to recover overall shape (e.g., fat or thin birds), it cannot capture fine details of birds with asymmetrical articulation or rare poses. \\
\indent As an extension of the CMR, the \textit{unsupervised CMR} (U-CMR) \cite{ref8} uses a single 3D template shape and a set of possible camera hypotheses (i.e., camera-multiplex) to address the requirement of keypoint labels and potential local minima problem for camera pose prediction. Inspired by particle filtering, the U-CMR utilizes a weakly-supervised iterative optimization approach to learn camera pose distribution for every shape and texture prediction. It comprises an encoder and two modules for predicting shape and texture, as well as threshold-based background subtraction and hole-filling to compute masks. The U-CMR can handle sharp long tails, protrusions of legs and beaks, and various bird types, although it fails to recover flying birds (large articulations) and birds with open wings or twisted heads due to its lack of an articulation model. \\
\indent Besides, the \textit{unsupervised mesh reconstruction} (UMR) \cite{ref9} proposes a self-supervised model (to address the keypoint-dependency of the CMR \cite{ref5}) that renders the semantic parts of each object to form a collection of deformable parts in both 2D and 3D space and their consistency (e.g., for wings on birds) across different instances of the similar category. It couples a collection of semantic parts (decomposed from a 2D image) with different object instances to build a canonical semantic UV texture map for each category, helping to create a template that captures major shape characteristics and semantic parts of the objects. Although the model is independent of any mesh or shape prior, the UMR involves some challenges such as poor segmentation method, rare camera pose instance and model limitation to capture certain shape characteristics details (e.g., the two wings of flying birds). 
For dealing with large shape variation and background distractors, the \textit{weakly-supervised CMR} (WCMR) \cite{ref33} utilizes a fusion module, shape-sensitive geometric constraints, and background manipulation to combine multi-scale features, supervise feature extraction \& shape reconstruction, and improve robustness, respectively. By exploiting multi-scale features as well as the side-output mask, adaptive edge, \& initial shape constraints, this method can cope with appearance variations, directly predict masks from the encoder's intermediate layers, be able to handle ambiguous shapes, and make explicit use of an initial shape. \\
\indent On the other hand, the mapping of 2D image pixels to the locations on an abstract 3D model of a category can be considered the key to a rich understanding of objects and correspondence inference. Hence, the \textit{canonical surface mapping} (CSM) \cite{ref29} completes the cycle of pixels to 3D to pixels for birds by combining this mapping with 3D projection as well as exploiting a geometric cycle consistency loss to bypass reliance on keypoint or pose supervision. To do so, a parameterized network with U-Net architecture learns to infer canonical surface mapping using an input image and foreground mask with the aid of an additional network for predicting camera parameters. While an additional per-pixel mask predictor generates the foreground mask and ignores background pixels for inferring, a weak perspective transformation and a \textit{neural mesh renderer} (NMR) \cite{NMR} are employed to model camera projection and render a depth map, respectively. Despite the promising results, the primary limitations of the CSM approach pertain to handling significant shape differences or large articulations of categories (e.g., for non-rigid objects).
As a solution to the CSM limitations, geometric consistency can be enforced without direct supervision to explicitly learn pose and articulation using input images and associated foreground mask labels. The \textit{articulated CSM} (ACSM) \cite{ref3} exploits this consistency and auxiliary losses to synchronize the pose, articulation, and canonical surface mapping and prevent degenerate or trivial solutions, respectively. It employs two ResNet-18-based encoder-decoder networks to jointly infer the mapping as well as the camera pose and articulation of 3D templates, while it can be extended to scenarios with available annotations. However, this method is restricted to intrinsic shape variations of species and results in sub-optimal performance for birds due to its ambiguities for their parts (e.g., beak and wing). \\
\indent Despite the approaches that separate modules of a network regress camera pose and non-rigid deformation, the \textit{To-The-Point} \cite{ref28} (TTP) jointly learns them by a lightweight per-sample differentiable optimization given 2D images, foreground masks, and optional keypoints. This method uses a sampling-based texture approach and as-rigid-as-possible (ARAP) constraint \cite{ARAP} to estimate accurate correspondence predictions and constraint arbitrary deformations (e.g., anomalies). The TTP uses Mask-RCNN and ResNet-based encoder to obtain foreground masks and map an image to latent feature maps, respectively. Although the mesh can be efficiently recovered in a few iterations (or even one), inaccurate camera poses and incorrect 2D points can negatively impact its performance (e.g., for flying birds). \\
\indent The learning of explicit 3D representations entails limitations regarding scalability beyond a few categories as well as coping with intra-instance shape variations such as thin and fat birds. However, inferring the implicit shape and texture of a 3D object generally relies on 3D supervision and templates. The \textit{implicit mesh reconstruction} (IMR) \cite{ref36} addresses these restrictions through learning 3D inference with only in-the-wild image collections along with rough object segmentation. The IMR uses geometric consistency to bypass the need for direct supervision, thereby promoting implicit shape and texture representations. After obtaining an initial 3D mesh from an encoder, this method parameterizes the shape and texture to learn the category-level latent representations. Besides, it uses a simple four-layer feedforward MLP to instantiate the shape and texture implicit functions. 
Although this method addresses several bottlenecks and provides good prediction of shape and texture for different object categories, it is limited to the rendering of objects that are not subject to large shape variations. In addition, the IMR assumes that the object is not occluded, which may affect the overall performance. \\
\indent Despite relying on weakly- or self-supervised approaches, trained networks may predict unstable and inconsistent shapes for video sequences because of perturbations over time. The \textit{asymmetric CMR} (ACMR) trains an image-based network on a collection of category-specific images with an ARAP constraint \cite{ARAP} to predict the shape, texture, and camera pose and avoid the symmetric assumption to effectively handle scenarios such as birds tilting/rotating their heads or deforming from standing to flying. This method uses a weighted combination of learned shape bases to recover asymmetric shapes and prevent large motion deformations of shape models, which can be trained by an encoder using either object silhouettes \& 2D keypoints or only object silhouettes.
The ACMR also can exploit online adaptation to generalize the learned model to input videos and self-supervised setting. For online adaptation, all parameters of the learned reconstruction model are fine-tuned, and then random parts onto the middle frame are painted and propagated backward to the first frame and forward to the last frame. During self-supervised training, this method stabilizes the predictions by taking shape symmetry into account, as opposed to its weakly-supervised setting that benefits from keypoint supervision. 
Lastly, to deform the generic template model to new species and the individual instances sequentially, the \textit{Avian species-specific} (AVES) \cite{ref4} extends the articulated bird mesh obtained in \cite{ref2} considering lack of strong deformable shape prior. It aligns a generic template model with the samples and then deforms it to new species to acquire the mean shape of them. The mean shape is further deformed to capture variations among different instances of the new species using a combination of shape basis.
\vspace{-.1cm}
\subsection{Video Frames-based Approaches}
The dense correspondences between consecutive frames can be leveraged as the self-supervision to reconstruct the 3D shapes of articulated objects. The \textit{articulated categories from motion} (ACM) \cite{ref31} uses this supervision to optimize regularized mesh deformations using per-frame semantic segmentation masks and keypoint reprojections. This method exploits single-frame networks (i.e., one ResNet-based encoder and three decoders for predictions) that jointly infer category objects' 3D shape, camera pose, and texture. It can correctly capture the wing variation of birds (highly flexible wings), which is challenging for alternative methods. However, incorrect global camera parameters (e.g., scale) can cause failures caused by the small diversity of appearances in videos. \\
\indent To avoid relying on category-specific 3D shape templates, the \textit{learning articulated shape reconstruction} (LASR) \cite{ref26} jointly learns 3D shape, articulation, and camera parameters from a single monocular video to provide a template-free approach. The joint recovery mechanism is obtained by solving inverse graphics problem via gradient-based optimization. It employs a network to parameterize the camera parameters and articulation. Because LASR relies on generic shape and motion priors, it can be applied to a wider range of non-rigid shapes. However, the main drawback of LASR is failure to render invisible surface of occluded objects that are missed by mask annotations. \\
\indent Finally, the \textit{deformable objects from videos} (DOVE) \cite{ref34} method employs video frames from short clips, corresponding 2D masks, and optical flows to reconstruct an articulated 3D shape with texture and rigid pose of an object instance, without requiring category-specific 3D template. Adopting a photo-geometric auto-encoding, this method takes a base mesh to optimize its vertices by three networks (rigid pose, shape, \& texture), whose recombined information provides supervision. Despite the simple geometry and deformation structure of birds, the performance of this method can be degraded when the birds are away from the camera or have pose variations of more than 90 degrees. In addition, the results may be affected by the required segmentation masks, motion blur, or missing fine details (e.g., legs and beak).

\section{Conclusion}
In this paper, we presented recent advances to the learning-based 3D reconstruction of bird species using single-view images and videos. The challenging idea of recovering the pose and shape of birds, which can play an important role in understanding the collective behavior of species in the wild, has inspired a new research direction in the area of computational ecology and neuroscience. Although each approach in this area shows superior results under certain challenging scenarios, learning schemes, and geometric variations, we acknowledge that there are still several challenges that exist to solving this highly ill-posed problem. However, we believe that highlighting the strengths and weaknesses of each 3D reconstruction method for flying birds provides great value for future studies in this research direction.

\section*{Acknowledgments}
\noindent This work was supported in-part by the fRI Research-Mountain Pine Beetle Ecology Program.

\bibliographystyle{IEEEtran}
\bibliography{0_ref}

\end{document}